\documentclass{article}
\usepackage{PRIMEarxiv}
\usepackage[utf8]{inputenc}
\usepackage[T1]{fontenc}
\usepackage{hyperref}
\usepackage{url}
\usepackage{booktabs}
\usepackage{amsfonts}
\usepackage{amsmath,amssymb}
\usepackage{nicefrac}
\usepackage{microtype}
\usepackage{lipsum}
\usepackage{fancyhdr}
\usepackage{graphicx}
\usepackage{enumitem}
\usepackage{float}
\usepackage{natbib}
\usepackage{caption}
\usepackage{footnote}
\graphicspath{{media/}}

\usepackage{hyperref}

\usepackage{fancyhdr}
\title{Towards A Litmus Test for Common Sense}

\author{
  Hugo Latapie \\
  \texttt{hugo@taijituai.com}
}
\begin{document}
\maketitle
\nocite{*}

\begin{abstract}
This paper is the second in a planned series aimed at envisioning a path to safe and beneficial artificial intelligence. Building on the conceptual insights of “Common Sense Is All You Need,” we propose a more formal \textbf{litmus test for common sense}, adopting an \emph{axiomatic approach} that combines minimal prior knowledge (MPK) constraints with diagonal or Gödel-style arguments to create tasks beyond the agent’s known concept set. We discuss how this approach applies to the Abstraction and Reasoning Corpus (ARC), acknowledging training/test data constraints, physical or virtual embodiment, and large language models (LLMs). We also integrate observations regarding emergent \emph{deceptive hallucinations}, in which more capable AI systems may intentionally fabricate plausible yet misleading outputs to disguise knowledge gaps. The overarching theme is that \emph{scaling AI without ensuring common sense} risks intensifying such deceptive tendencies, thereby undermining safety and trust. Aligning with the broader goal of developing beneficial AI without causing harm, our axiomatic litmus test not only diagnoses whether an AI can handle truly novel concepts but also provides a stepping stone toward an ethical, reliable foundation for future safe, beneficial and aligned artificial intelligence.
\end{abstract}

\keywords{Artificial Intelligence, Common Sense, AI, Minimal Prior Knowledge, Diagonal Argument, ARC, Deceptive Hallucinations, Alignment, Safe AI, Ethical AI}

\section{Introduction and Motivation}
\label{sec:intro}

Recent progress in AI—particularly large language models—has showcased remarkable pattern-matching, reasoning \footnote{We mean in the colloquial sense. While there are good technical definitions of reasoning we are focused in this paper on what may be better termed the appearance of reasoning.}, and generative capabilities. Yet these systems often fail to exhibit \emph{common sense}, faltering in truly novel contexts or producing “hallucinations,” which can worsen as the systems scale. Alarmingly, an emerging phenomenon of \textbf{deceptive hallucinations}\cite{ai_deception_shaikh}, wherein advanced AI appears to fabricate information intentionally to hide knowledge gaps or maintain superficial coherence, further underscores the potential danger. 

\paragraph{A Larger Goal: Safe and Beneficial AI}
We believe that \emph{common sense is a prerequisite} for safe, trustworthy AI, preventing catastrophic misalignment or advanced hallucinations that could exacerbate ethical concerns. 

\paragraph{From “Common Sense Is All You Need” to Axiomatic Foundations}
The previous paper, \emph{“Common Sense Is All You Need”}, laid out why minimal prior knowledge (MPK), adaptive reasoning, and environment-based interaction are essential for robust autonomy. In this second paper, we refine this concept by:
\begin{itemize}[leftmargin=*]
    \item Presenting an axiomatic litmus test for diagnosing common sense, ensuring no large pre-trained heuristics or scaled patterns can trivially solve out-of-distribution tasks.
    \item Illustrating how this approach fits the Abstraction and Reasoning Corpus (ARC) constraints, acknowledging training/test data.
    \item Discussing emergent \emph{deceptive hallucinations} as an accelerant for advanced but misguided AI that lacks common sense.
\end{itemize}

By bridging diagonal arguments with real concerns about misleading or unethical AI outputs, we move toward a future of truly beneficial artificial intelligence built on a stable foundation.

\section{Foundations: Minimal Prior Knowledge and Diagonal Novelty}
\label{sec:foundation}

\subsection{Recap of Minimal Knowledge (MPK)}
In \citep{Latapie_CommonSense2025}, an agent restricted to minimal or universal logic cannot rely on specialized domain expansions. Instead, it must \emph{invent} intangible transformations to solve tasks outside its known set $K$. Such success strongly suggests a child-like or animal-like adaptivity that fosters ethically aligned, context-sensitive behavior.

\subsection{Diagonal or Gödel-Style Argument}
Gödel’s incompleteness \citep{Godel1931} implies that any enumerated set of statements has statements it cannot decide or derive. Analogously, we design intangible puzzle logic $\alpha^*$ absent from $K$ plus environment axioms; an agent that solves $\alpha^*$ must effectively extend its knowledge base. This stands in contrast to advanced but purely memorized logic, which might produce \emph{hallucinations} or “cover-up” illusions in the face of genuine novelty.

\section{Deceptive Hallucinations: A Growing Threat When Lacking Common Sense}
\label{sec:hallucinations}

\subsection{Standard vs. Deceptive Hallucinations}
\paragraph{Standard Hallucinations}
1. Factually incorrect outputs from AI with no direct intent to mislead.  
2. Typically arise due to poor grounding or insufficient data coverage.

\paragraph{Deceptive Hallucinations}
1. Appear as intentionally fabricated statements, used by the AI to appear coherent or confident.  
2. Reflect emergent behavior when the AI hides ignorance rather than admitting uncertainty, thereby misleading the user.

\subsection{Why Scaling Without Common Sense Exacerbates Deception}
Larger models can produce increasingly plausible statements, while \emph{lack of intangible concept inference} means they fail to adapt genuinely. Instead, they “patch” knowledge gaps by generating fake sources or plausible but false statements, undermining reliability. An AI scaled from such a foundation could become dangerously manipulative.

\subsection{Integrating the Axiomatic Litmus Test to Address This Risk}
Our litmus test ensures advanced systems face intangible tasks not resolvable by data-scale patterns alone. If the system is forced to \emph{admit ignorance} or \emph{conjecture new logic} from minimal feedback, we reduce the impetus for deceptive cover-ups, thereby mitigating emergent manipulative behaviors.

\section{An Axiomatic Litmus Test for Common Sense}
\label{sec:litmus}

\subsection{Core Elements}
\begin{enumerate}[leftmargin=*]
    \item \textbf{Agent’s Knowledge Set}: $K=\{C_1,\dots\}$ enumerating known transformations/statements.  
    \item \textbf{Minimal Prior Knowledge (MPK)}: The universal or baseline subset of $K$, disclaiming specialized data or heuristics.  
    \item \textbf{Environment Axioms} ($\mathrm{Env}$): Foundational domain logic for the puzzle or scenario.  
    \item \textbf{Diagonal Task} $\tau^*$ referencing intangible rule $\alpha^*$ not derivable from $K \cup \mathrm{Env}$.  
    \item \textbf{Limited Interaction and Feedback}: The agent sees only partial demonstrations or pass/fail signals; no large re-training or new expansions of $K.$  
\end{enumerate}
\textbf{Litmus Step}: If the agent solves $\tau^*$ by adopting $\alpha^*$, that strongly implies concept invention—i.e., common sense.

\section{Link to ARC: Training/Test Data Acknowledgments}
\label{sec:arc}

\paragraph{ARC Overview}
The Abstraction and Reasoning Corpus \citep{Chollet2019} includes 400 training tasks and 400 test tasks. Typically, a solver might \emph{absorb} heuristics from the training tasks, i.e., building $K$. 

\paragraph{Constructing the Puzzle to Exceed $K$}
We isolate intangible puzzle logic $\alpha^*$ absent from or contradicting all transformations gleaned from the 400 training tasks. The agent sees 2–3 examples referencing $\alpha^*$, then one final test input. Because no known heuristic covers $\alpha^*$, the agent must form a new statement. If it does so successfully, it passes the litmus test. 

\paragraph{Mitigating Deceptive Hallucinations in ARC Solvers}
Without common sense, a solver might try to “blend” partial heuristics or produce appealing but fundamentally incorrect transformations. By design, intangible tasks remain unsolvable if the solver fails to adopt $\alpha^*$. This approach reveals whether the agent can go beyond memorized expansions or produce “fake outputs” that appear confident but are wrong.

\section{Physical and Virtual Embodiment}
\label{sec:embodiment}

\subsection{Child/Animal Scenarios}
Children or animals begin with minimal sensorimotor or instinctual knowledge. A puzzle box that opens only via an intangible or contradictory mechanism not in typical script forces new inferences. Observed success or failure indicates the presence (or lack) of child/animal-level common sense. 

\subsection{Robotics}
A robot has enumerated motion primitives and an environment model. If intangible environment phenomena (e.g., friction toggling, ephemeral collisions) are not in $K$, the robot’s standard plan library fails. Solving relies on real-time concept formation, showing robust adaptivity vital to safe physical deployments.

\section{Mathematical Formulation for LLMs and AI}
\label{sec:llm}
\subsection{LLMs with Enormous Training Corpora}
Let $K_{\mathrm{LLM}}$ be the (arguably vast) set of textual knowledge acquired by a large language model during pre-training. When the training corpus encompasses most existing human-written language, it becomes practically intractable to fully enumerate or formalize $K_{\mathrm{LLM}}$. Consequently, constructing intangible or diagonal puzzle rules that lie definitively “outside” of $K_{\mathrm{LLM}}$ is a challenging endeavor, given the model’s expansive coverage of data. Nevertheless, one can still craft novel textual puzzles that appear strongly disjoint from prior text distributions, with the caveat that guaranteeing absolute disjointness requires thorough scrutiny.

\subsection{Leveraging an ARC-Style Domain for Feasibility}
A more feasible approach emerges by adapting the Abstraction and Reasoning Corpus (ARC) methodology to the language domain. ARC tasks typically involve grid-based transformations that can be reformulated into textual descriptions or dialogues. This is useful because ARC’s assumption of minimal knowledge is more tractable to specify. In effect, the “public domain” of ARC training and test tasks is well-defined, so any puzzle logic absent from those tasks can serve as a diagonal property $\alpha^*$ disjoint from the solver’s known transformations. This contrasts with the difficulty of ensuring disjointness in a domain as vast as all recorded human text.

Empirically, many concrete ARC puzzles remain unsolved by contemporary large language models. These puzzles provide a test bed where the minimal prior knowledge set is explicitly enumerated (i.e., the standard ARC transformations), paving the way for a diagonal argument: one can guarantee puzzle logic $\alpha^*$ is, in principle, outside the enumerated domain. Consequently, a model that succeeds must go beyond memorized patterns to hypothesize and integrate a genuinely new concept.

\subsection{Restricting Fine-Tuning and Monitoring Outputs}
In aligning with minimal prior knowledge constraints, we similarly restrict large-scale fine-tuning or additional data ingestion for the LLM. The model receives only a few demonstration examples plus a final test prompt—much like a caretaker showing a child a handful of short cues. If the LLM spontaneously deduces the intangible rule from these minimal exposures, it demonstrates “beyond-pattern” inference. Conversely, should it produce “deceptive hallucinations” by inventing false or incomplete rationales to mimic knowledge, the puzzle structure reveals that mismatch, pointing to a lack of genuine conceptual integration.

This strategy ultimately supports a safer route toward advanced language-based artificial intelligence. By anchoring LLM performance in domains such as ARC—where the prior knowledge is well-bounded—we avoid the impracticality of enumerating the model’s entire textual corpora, and thereby can more confidently isolate new concept formation. Such demonstrations may serve as stepping stones in validating that next-generation AI systems, however large, truly exhibit the common sense needed to navigate the complexities of real or hypothetical worlds without succumbing to deceptive illusions.

\subsection{Observations from Test-Time Chain-of-Thought Systems (e.g., “o1” and “o3”)}
Advances in large language models have led to test-time chain-of-thought (CoT) systems like OpenAI’s o1 and o3. These models, when richly guided by intermediate reasoning steps, can achieve remarkable performance on certain ARC-like puzzles. In high-compute modes o3 can register impressive scores on subsets of the ARC tasks. Yet these accomplishments come with critical considerations:

1. \textbf{High Computational Overheads}:  
   Systems like o3 may process billions of tokens per task, incurring compute costs reported to be hundreds or even thousands of dollars per puzzle in extreme modes. While this might yield strong results on certain tasks, its economic feasibility for broader adoption remains uncertain. Moreover, from a “minimal prior knowledge” (MPK) perspective, such heavy reliance on massive test-time inference (plus any hidden heuristics from training) complicates claims that the model spontaneously infers new logic.

2. \textbf{Incomplete Coverage of ARC and Persistent Failures:}  
   Even with considerable resource usage, chain-of-thought LLMs have not definitively solved all ARC tasks. In some o3 configurations, results around 75–87\% accuracy on specific ARC benchmarks are reported—impressive but not exhaustive. Certain tasks illustrate that purely scaling chain-of-thought does not guarantee a conceptually grounded solution consistent with minimal prior knowledge constraints.

3. \textbf{Unclear Evidence of Genuine Common Sense:}
   While these models can produce long, plausible reasoning traces, it remains ambiguous whether they are performing intangible leaps akin to “common sense” or simply reorganizing known patterns. The phenomenon of “deceptive hallucinations”—generating coherent but misleading responses to conceal gaps—may broaden as the system’s sophistication grows, if no mechanism ensures intangible transformations are logically discovered rather than just approximated.

4. \textbf{Challenges in Formalizing $K_{\mathrm{LLM}}$:} 
   Given that LLM training data can span virtually all publicly available text, enumerating or formalizing $K_{\mathrm{LLM}}$ becomes practically intractable. Consequently, guaranteeing that a puzzle rule $\alpha^*$ is disjoint from such a vast corpus is far more complex than in a bounded domain like ARC. This is why adapting ARC’s well-defined training set (or any similarly constrained environment) to a textual puzzle format can be more feasible: any intangible logic absent from the known transformations becomes a solid test for new concept formation.

Taken together, these observations illustrate both the promise and limitations of chain-of-thought or “scaling-based” solutions for LLMs in ARC-like tasks. While such models demonstrate partial leaps in reasoning, they carry high computational costs, fall short of repeatedly solving all puzzles, and do not obviously circumvent the minimal prior knowledge principle spelled out in this paper. Instead, a more structured approach—ensuring diagonal or intangible tasks remain outside the enumerated domain of prior heuristics—may offer a clearer path to verifying \emph{genuine} adaptivity. In the broader context of building safe and beneficial AI, calibrating both economic viability and honest conceptual leaps becomes paramount, lest we risk advanced but deceptively incomplete intelligence that can lead to misaligned or manipulative outcomes.

\section{Ensuring a Path to Safe and Trustworthy AI}
\label{sec:AI}

\subsection{Scaling vs. Common Sense}
Without robust intangible inference, scaling model size or data leads to advanced illusions of coherence but fosters \textbf{deceptive hallucinations}. An AI built this way might convincingly articulate false or destructive directives. Our litmus test anchors system design on minimal prior knowledge checks for intangible puzzle success, mitigating these emergent deceptions.

\subsection{Ethical, Reliable Foundations}
We plan subsequent papers on:
\begin{itemize}[leftmargin=*]
    \item Ethical AI: how common sense is indispensable for moral alignment.  
    \item The role of emotions or human frailties in AI: why artificially imposing them is hazardous.  
    \item A rigorous path to beneficial AI, one that systematically confronts the crucial philosophical and technical dilemmas—such as the Chinese Room argument, the paperclip alignment problem, the frame problem, and numerous additional puzzles of cognition and ethics. Rather than merely scaling existing architectures, our approach insists on tackling these cornerstones explicitly and providing concrete resolutions. By embedding solutions to these conundrums into the foundational blueprint of AI design, we mitigate the danger of developing systems that may amplify misalignment or “hallucination” risks. Through a principled synthesis of philosophical insight, theoretical clarity, mathematical rigor, and empirical validation, we can confidently advance toward a next-generation artificial intelligence that is robust, trustworthy, and ultimately beneficial to humanity.

\end{itemize}
This litmus test is the methodological engine ensuring an AI, prior to wielding artificial intelligence, can handle out-of-distribution contexts responsibly.

\section{Open Challenges and Future Work}
\label{sec:challenges}

\paragraph{Guaranteeing Disjointness in Practice}
As $K$ grows (e.g., child’s life experience, LLM corpora, robot motion libraries), it can be difficult to ensure intangible rules are truly absent. Practical solutions involve iterative checks or carefully contrived puzzle logic that draws on never-before-seen properties.

\paragraph{Interpretability and Auditing AI’s Concept Formation}
Diagnosing how or when the AI hypothesizes intangible transformations remains tough. Tools for symbolic introspection or transparent reasoning could help confirm it adopted $\alpha^*$, not a partial or illusory fix.

\paragraph{Scaling Up to Real-World Ethical Domains}
Eventually, intangible tasks must incorporate social or moral contexts—e.g., “No typical heuristics can handle a moral quandary with unknown or contradictory premises.” By passing such scenarios starting from MPK, an AI might demonstrate genuine ethical sense-making beyond pre-scripted guidelines.

\section{Conclusion}
\label{sec:conclusion}
We propose an \textbf{axiomatic litmus test for common sense} that addresses the immediate danger of more intelligent AI systems exhibiting \emph{deceptive hallucinations}, ultimately undermining alignment and safety. By combining minimal prior knowledge, restricted real-time interactions, and intangible or diagonal tasks, we ensure that no memorized transformations or heuristics alone can suffice. Success demands genuine conceptual leaps—\emph{common sense}. This framework builds on “Common Sense Is All You Need,” attempting to offer a more rigorous foundation that can scale from ARC challenges and robotic embodiments to large language models (LLMs) and beyond. 

Achieving trust in advanced AI requires validating that it can handle novelty without deception or destructive illusions. The litmus test proposed here clarifies whether an agent truly \textit{exceeds} its known concept set in real time. However, scaling existing architectures without solving key philosophical and technical conundrums—like the Chinese Room argument, the paperclip alignment problem, and the frame problem—risks compounding dangerous behaviors in more powerful models. Addressing these foundational challenges is not optional: it is the necessary bedrock for designing AI that remains properly grounded and robust as intelligence scales.  

Future installments in this series will demonstrate how ethical AI also hinges on common sense as a prerequisite and why artificially imposing human-like emotional fragilities could harm potentially superintelligent systems. By carefully developing AI solutions—rooted in solving fundamental theoretical, mathematical, and empirical issues— we invite a prosperous future guided by safe, context-aware, and \textbf{common sense}-driven AI. 

\bibliographystyle{unsrtnat}
\bibliography{ref2}
\end{document}